\documentclass[letterpaper, 10 pt, journal, twoside]{IEEEtran}  

\IEEEoverridecommandlockouts                              

\usepackage[utf8]{inputenc}
\usepackage[T1]{fontenc}

\usepackage[pdftex]{graphicx}

\usepackage{cite}

\usepackage{xcolor}
\usepackage{hyperref}
\usepackage{booktabs}
\usepackage{threeparttable}

\usepackage{amsmath}

\DeclareMathOperator*{\argmin}{arg\,min}

\usepackage{amssymb}
\usepackage{algorithm}
\usepackage{algpseudocode}

\title{A Unified Bi-directional Model for Natural and Artificial Trust in Human--Robot Collaboration}

\author{Hebert~Azevedo-Sa$^{1}$,
        X.~Jessie~Yang$^{1,2}$,
        Lionel~P.~Robert~Jr.$^{1,3}$
        and~Dawn~M.~Tilbury$^{1,4}$
\thanks{Manuscript received: February, 23, 2021; Revised May, 12, 2021; Accepted June, 2, 2021.}
\thanks{This paper was recommended for publication by Editor Gentiane Venture upon evaluation of the Associate Editor and Reviewers' comments.} 
\thanks{This work was partially supported by the National Science Foundation and by the Brazilian Army's Department of Science and Technology.}
\thanks{$^{1}$Hebert~Azevedo-Sa, X.~Jessie~Yang, Lionel~P.~Robert~Jr. and Dawn~M.~Tilbury are with the Robotics Institute, University of Michigan, Ann Arbor,
MI, 48109 USA. {\tt \footnotesize \{azevedo, xijyang, lprobert, tilbury\}@umich.edu.}}
\thanks{$^{2}$X.~Jessie~Yang is with the Department of Industrial and Operations Engineering, University of Michigan, Ann Arbor.}
\thanks{$^{3}$Lionel~P.~Robert~Jr. is with the School of Information, University of Michigan, Ann Arbor.}
\thanks{$^{4}$Dawn~M.~Tilbury is with the Department of Mechanical Engineering, University of Michigan, Ann Arbor.}
\thanks{Digital Object Identifier (DOI): see top of this page.}
}

\begin{document}

\maketitle

\markboth{IEEE Robotics and Automation Letters. Preprint Version. Accepted June, 2021}
{Azevedo-Sa \MakeLowercase{\textit{et al.}}: Unified Bi-directional Model for Natural and Artificial Trust in Human--Robot Collaboration} 

\begin{abstract}
We introduce a novel capabilities-based bi-directional multi-task trust model that can be used for trust prediction from either a human or a robotic trustor agent.
Tasks are represented in terms of their capability requirements, while trustee agents are characterized by their individual capabilities.
Trustee agents' capabilities are not deterministic; they are represented by belief distributions.
For each task to be executed, a higher level of trust is assigned to trustee agents who have demonstrated that their capabilities exceed the task's requirements.
We report results of an online experiment with 284 participants, revealing that our model outperforms existing models for multi-task trust prediction from a human trustor.
We also present simulations of the model for determining trust from a robotic trustor.
Our model is useful for control authority allocation applications that involve human--robot teams.
\end{abstract}


\section{Introduction}
\label{sec:intro}

\IEEEPARstart{W}{ould} you \textit{trust} someone to drive you in an overcrowded city with heavy traffic?
You probably would, if you knew that person was a capable driver.
Most certainly, to ultimately gain your trust, the potential \textit{trustee} driver must demonstrate her/his competence by providing you---the \textit{trustor} passenger---with a positive experience.

The driver--passenger example is only one of countless situations involving trust between two agents in a trust relationship: the trustor (the one who trusts) and the trustee (the one to be trusted).
Trust pervades our relationship with other people, with organizations, and with machines \cite{barber1983logic, Mayer1995AnTrust, muir1994trust, lee2004trust}.
Trust depends on both the trustor's and the trustee's characteristics and is revealed when the trustor takes the risk of being vulnerable to the trustee's actions \cite{Mayer1995AnTrust}.

Human--robot interaction (HRI) researchers have proposed predictive trust models that try to capture how a human trustor develops trust in a robotic trustee \cite{Xu2015OPTIMo:Collaborations, Soh2020Multi-taskInteraction, you2018human}.
A perspective that is generally overlooked, however, is how trust from a robotic trustor should develop over interactions with a trustee.
In this work we distinguish between human trust, which we label as \textit{natural trust}, from robotic trust, which we label as \textit{artificial trust}.
Current trust models are focused on natural trust and are useful for trust-aware decision-making, which requires the robot to estimate the human's trust in the robot to plan actions in an HRI setting.

\begin{figure}
  \centering
  \includegraphics[width=1.0\linewidth]{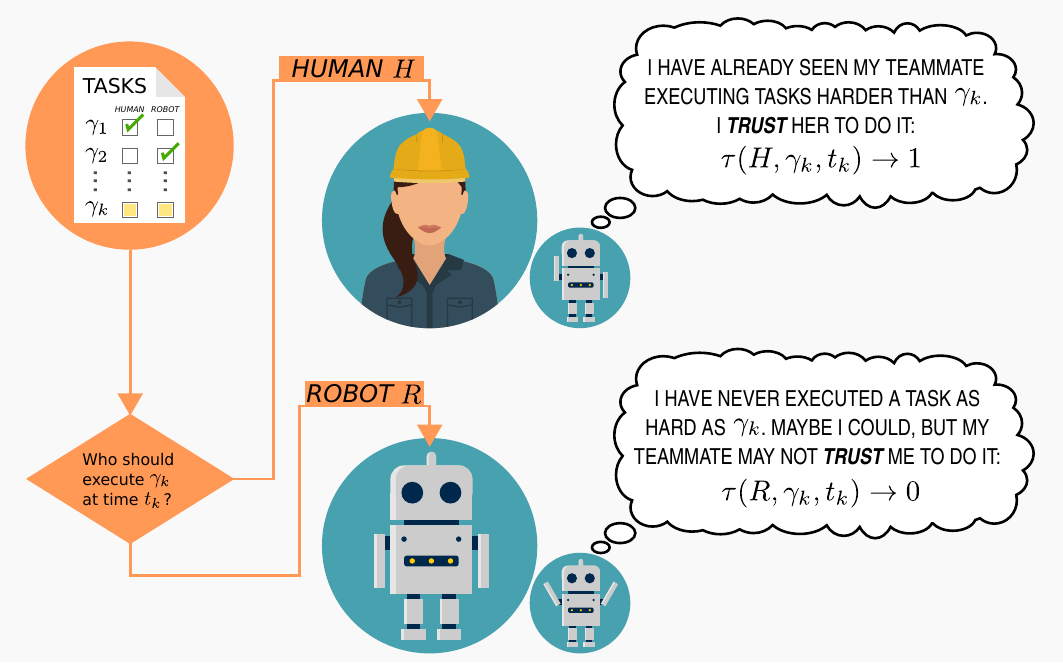}
  \caption{A team formed by human $H$ and a robot $R$ that must collaborate sequentially executing tasks.
  Each task must be executed by one of the agents.
  The bi-directional trust model can be used for predicting a human's  trust in a robot to execute a task, and to predict how much humans can be trusted to execute a task.}
  \label{fig:abstract_fig}
\end{figure}

Existing trust models have several shortcomings that hinder their ability to predict humans' natural trust and limit their application for robots' artificial trust computation.
First, current trust models are limited in their ability to characterize the tasks that should be executed by trustees.
Tasks must be characterized in terms of what capabilities and which proficiency levels (in those capabilities) are required from trustees to execute those tasks.
For instance, driving requires certain levels of cognitive, sensory and physical capabilities from drivers \cite{Anstey2005CognitiveAdults}.
Second, current trust models fall short of describing the trustee agents in terms of their proven capabilities.
Trustees' capabilities characterization and quantification are important because, when a trustor knows that the trustee is or is not capable of meeting the task requirements, the trustor's trust in the trustee to execute that task is higher (or lower).
Finally, because of a lack of trustee capability characterization, current trust models are applicable for natural trust, or understanding human trust in a robot, but not for artificial trust, especially for determining how a robot should trust a human.
Existing models are performance-centric and ignore \textit{non-performance trustees' capabilities} or \textit{factors}, which are needed for determining artificial trust.
To accommodate both natural and artificial trust in (human or robotic) trustees, a computational model of trust must be able to  consider assessments of a trustee's non-performance capabilities, such as honesty, benevolence or integrity levels \cite{Mayer1995AnTrust, malle2021multidimensional}.
Therefore, although existing trust models are sufficient for planning algorithms, these trust models can not be used in more sophisticated control authority allocation applications, which are likely to be based on comparisons between the human’s trust in the robot and the robot’s trust in the human \cite{azevedo2021handling}.

To address those shortcomings, we propose a novel capabilities-based bi-directional trust model.
Our model characterizes tasks on a set of standard requirements \textcolor{black}{that can represent either performance or non-performance capabilities that affect trust}, and builds trustee capability profiles based on the trustee's \textcolor{black}{history on executing those tasks}.
Trust is represented by the probability that an agent can successfully execute a task, considering that agent's capability profile (built after observations).
\textcolor{black}{By considering the agent's capabilities (performance or non-performance) \cite{malle2021multidimensional} and the task requirements, our model can be used to determine a robot's artificial trust in a trustee agent.}
Moreover, our model can be used for predicting trust transfer between tasks, similar to the model proposed in \cite{Soh2020Multi-taskInteraction}.
However, as compared to \cite{Soh2020Multi-taskInteraction}, our model improves trust transfer predictions by representing tasks in terms of capability requirements instead of using natural language processing (NLP) similarity metrics.
We show the superiority of our trust model by comparing its prediction results with those from other models, using a dataset collected in an online experiment with 284 participants.
In sum, our contributions with this work are:

\begin{itemize}

    \item a new trust model that (i) can be used for the \textit{artificial} trust computation and (ii) outperforms existing models for multi-task \textit{natural} trust transfer prediction; and
    
    \item an online experiment that resulted in a dataset relating trust and task capabilities measurements.

\end{itemize}

\section{Trust in Human--Robot Interaction}
\label{sec:background}

\subsection{Origins and Current Stage of Trust in HRI}


Trust in robots that interact with humans can be considered as an evolution of trust in automation, which in turn has evolved from theoretical frameworks on interpersonal trust.
Muir \cite{muir1987trust} proposed the concept of trust in automation after adapting sociologist interpersonal trust definitions \cite{barber1983logic, rempel1985trust} to humans and automated machines \cite{lee1992trust}.
Trust in automation is a dynamic construct \cite{Lewis2018TheInteraction} that can be directly measured with subjective scales \cite{muir1994trust, jian2000foundations} or can also be estimated through behavioral variables \cite{lee1994trust, Azevedo-Sa2020Real-TimeSystems}.

People's trust in an automated system must be calibrated, which means it has to align with the system's capabilities.
Miscalibrated trust is likely to lead to the inappropriate use of the system \cite{Lee2004, Lewis2018TheInteraction, Kok2020TrustOpportunities, Azevedo-Sa2020Context-AdaptiveVehicles}.
However, the evolution of automated systems into autonomous robots with powerful sensing technologies has paved the way for new trust calibration strategies.
Robots can now perceive and process humans' trust and take action to increase or decrease humans' trust when necessary \cite{Azevedo-Sa2020Context-AdaptiveVehicles, Chen2020Trust-AwareCollaboration, sheng2021trust-based}.

\subsection{Trust Definition}

Several trust definitions have been proposed, generally pointing to the trustor's attitude or willingness to be vulnerable to the trustee's actions \cite{lee2004trust, Mayer1995AnTrust}.
\textcolor{black}{In this work, we assume the (adapted) definition for trust recently proposed by Kok and Soh, which states that: ``given a trustor agent $A$ and a trustee agent $B$, $A$'s trust in $B$ is a multidimensional latent variable that mediates the relationship between events in the past and $A$'s subsequent choice of relying on $B$ in an uncertain environment'' \cite{Kok2020TrustOpportunities}.
Kok and Soh's definition establishes important aspects of our model, such as the multidimensionality of trust and its dependence on a history of events involving the trustor and the trustee agents.}

\subsection{Trust Computational Models}

Trust models are usually applied to determine how much a human trusts a robot to perform a task (e.g. Fig. \ref{fig:abstract_fig}, where the robot $R$ is chosen to execute a task).
The robot uses this estimate of human trust to predict the human's behavior, such as intervening on the task execution.
For example, trust models have been used in different trust-aware POMDP-based algorithms proposed for robotic planning and decision-making \cite{chen2018planning, sheng2021trust-based}.
Their objective is to eventually improve the robot’s collaboration with the human, using human trust as a vital factor when planning the robot's actions.

Planning and decision-making frameworks usually rely on the use of probabilistic models for trust \cite{Xu2015OPTIMo:Collaborations, Guo2020ModelingApproach, FooladiMahani2020ATeams}.
Xu and Dudek proposed an online probabilistic trust inference model for human--robot collaborations (OPTIMo) that uses a dynamic Bayesian network (DBN) combined with a linear Gaussian model and recursively reduces the uncertainty around the human operator's trust.
OPTIMo was tested in a human--unmanned aerial vehicle (UAV) collaboration setting \cite{Xu2015OPTIMo:Collaborations} and, although some dynamic models had been proposed before \cite{lee1992trust, desai2013impact}, OPTIMo was the first trust model capable of tracking human's trust in a robot with low latency and relatively high accuracy.
The UAV, with OPTIMo, was able to track the human operator's trust by observing how much the human intervened in the UAV's operation.

Other Bayesian models have been proposed since OPTIMo.
These models include personalized trust models that apply inference over a history of robot performances, such as \cite{FooladiMahani2020ATeams} and \cite{Guo2020ModelingApproach}.
Mahani et al. proposed a model for trust in a swarm of UAVs, establishing a baseline for human--multi-robot interaction trust prediction \cite{FooladiMahani2020ATeams}.
Guo and Yang \cite{Guo2020ModelingApproach} have improved trust prediction accuracy (as compared to Lee's ARMAV model \cite{lee1992trust} and OPTIMo \cite{Xu2015OPTIMo:Collaborations}) by proposing a formulation that describes trust in terms of Beta probability distributions and aligns the inference processes with trust formation and evolution processes \cite{Guo2020ModelingApproach}.
\textcolor{black}{Without explicitly modeling trust, Lee et al. showed that a robot that estimates and calibrates humans' intents and capabilities while making decisions can engender higher trust from humans \cite{Lee2020GettingCollaboration}.}

Although all previously mentioned approaches for trust modeling represent important advances in how we understand and describe humans' trust in robots, they suffer from a common limitation.
Those models depend on the history of robots' performances on unique specific tasks and are not applicable for trust transfer between different tasks. 
The issue of multi-task trust transfer was recently approached by Soh et al. \cite{Soh2020Multi-taskInteraction}, who proposed Gaussian processes and neural methods for predicting the transferred trust among different tasks that were described with NLP-based text embeddings.
A major goal for our model was to deepen that discussion and improve prediction accuracy for multi-task trust transfer by (i) describing tasks in terms of capability requirements, and (ii) describing potential trustee agents in terms of their proven capabilities that can be used to transfer trust to another task.

\textcolor{black}{The other major goal for our model was to be bi-directional, i.e., to be able to represent either natural trust or artificial trust.
Because the existing trust models are usually performance-centric, they are suited to represent humans' natural trust in robots.
Although mutual trust has been modeled as a single variable that depends on both the human's and the robot's performances on collaborative tasks \cite{wang2014human}, to represent a robot's artificial trust in humans, trust models must be more comprehensive.
Computational models of trust must consider not only performance factors but also non-performance factors that describe human trustees \cite{malle2021multidimensional, Mayer1995AnTrust, patacchiola2016developmental, Vinanzi2019WouldMind}.
Until recently, only a few trust models have considered the robot's trust perspective, focusing only on non-performance factors that affect trust.
For instance, a model that reproduces theory of mind (ToM) aspects in robots to identify deceptive humans has been proposed and applied in \cite{patacchiola2016developmental} and \cite{Vinanzi2019WouldMind}.
Our model is applicable for either natural or artificial trust because it explicitly considers a general form of agents' capabilities and task requirements, which can represent performance or non-performance trustee capabilities.}

\section{Bi-Directional Trust Model Development}
\label{ssec:problem_model}

\subsection{Context Description}

Consider the following situation: two agents (human $H$ or robot $R$) collaborate and must execute a sequence of tasks.
These tasks are indivisible and must be executed by only one agent.
The execution of each task can either succeed or fail.
For each task, one of the agents is in the position of trustor, and the other is the trustee.
Therefore, the trustor is vulnerable to the trustee's performance in that task.
From previous experiences with the trustee, the trustor has some implicit knowledge about the trustee's capabilities.
This implicit knowledge is used by the trustor \textcolor{black}{to} assess how likely the trustee is to succeed or fail in the execution of a task.
We define the terms and concepts we need for developing our trust model:

\textbf{Definition 1 - Task.} A \textit{task} that must be executed is represented by $\gamma \in \Gamma$. $\Gamma$ represents the set of all tasks that can be executed by the agents.

\textbf{Definition 2 - Agent.} An \textit{agent} $a \in \{H, R\}$ represents a trustee that could execute a task $\gamma$.

\textbf{Definition 3 - Capability.} The representation of a specific skill that agents have/that is required for the execution of tasks.
We represent a capability as an element of a closed interval $\Lambda_i = [0, 1]$, $i \in \{1, 2, 3, ..., n\}$, with $n$ being a finite number of dimensions characterizing distinct capabilities.

\textbf{Definition 4 - Capability Hypercube.} The compact set representation of $n$ distinct capabilities, given by the Cartesian product $\Lambda = \prod_{i=1}^n \Lambda_i = [0, 1]^n$.
This definition is inspired by the particular capabilities from Mayer et al.'s model \cite{Mayer1995AnTrust}, namely ability, benevolence and integrity, but the definition is intended to be broader than these three dimensions.

\textbf{Definition 5 - Agent's Capability Transform.} The agent capability transform $\xi: \{H, R\} \to \Lambda$ maps an agent into a point in the capability hypercube representing the \textit{agent's capabilities}, given by $\xi(a) = \lambda = (\lambda_1, \lambda_2, ..., \lambda_n) \in \Lambda$.

\textbf{Definition 6 - Task Requirements Transform.} The \textit{task requirements transform} $\varrho: \Gamma \to \Lambda$ maps a task $\gamma$ into the minimum required capabilities for the execution of $\gamma$, given by $\varrho(\gamma) = \bar{\lambda} = (\bar{\lambda}_1, \bar{\lambda}_2, ..., \bar{\lambda}_n) \in \Lambda$.


\textbf{Definition 7 - Time Index.} The \textit{time} is discrete and represented by $t \in \mathbb{N}$.

\textbf{Definition 8 - Task Outcome.} The outcome of a task $\gamma$ after being executed by the agent $a$ at the time $t$ is represented by $\Omega(\xi(a), \varrho(\gamma), t) \in \{0, 1\}$, where $0$ represents a failure and $1$ represents a success.
We also define the Boolean complement of $\Omega$, denoted by $\mho$, such that $\mho = 1$ when $\Omega = 0$, and $\mho = 0$ when $\Omega = 1$.

Leveraging the previous definitions, we can finally define trust.

\textbf{Definition 9 - Trust.} A trustor agent's trust in a trustee agent $a$ to execute a task $\gamma$ at a time instance $t$ can be represented by
\begin{equation}
\label{eq:trust}
\begin{aligned}
    \tau(a, \gamma, t) &= P\big(\Omega(\xi(a), \varrho(\gamma), t) = 1\big) \\
    &= 
    \int_{\Lambda} p\big(\Omega(\lambda, \bar{\lambda}, t)  = 1 | \lambda, t\big) bel(\lambda, t-1) d \lambda ,
\end{aligned}
\end{equation}
where $\lambda = \xi(a)$, $\bar{\lambda} = \varrho(\gamma)$, and $bel(\lambda, t-1)$ represents the trustor's belief in the agent's capabilities $\lambda$ at time $t-1$ (i.e., before the actual task execution). 
The belief is a dynamic probability distribution over the capability hypercube $\Lambda$.
Note that, at each time instance $t$, trust is a function of the task requirements $\bar{\lambda}$, representing a \textit{probability of success}  in $[0, 1]$.

\subsection{Bi-directional Trust Model}
\label{ssec:btm}

Our bi-directional model is defined by Eq. (\ref{eq:trust}) and depends on the combination of:
\begin{itemize}
\item a function to represent the ``trust given the trustee's capability'', represented by the conditional probability $p\big(\Omega(\lambda, \bar{\lambda}, t)  = 1 | \lambda, t\big)$; and 

\item a process to dynamically update the trustor's belief over the trustee capabilities $bel(\lambda, t)$.
\end{itemize}


We assume that an agent that successfully performs a task is more likely to be successful on less demanding tasks.
Conversely, an agent that fails on a task is more likely to fail on more demanding tasks.
We adapt the sigmoid function to represent that logic, and for each capability dimension we can write
\begin{equation}
\label{eq:trustGivenCapability_dim}
\begin{aligned}
    \tau_i = \Bigg[ \frac{1}{1 + e^{\beta_i (\bar{\lambda}_i - \lambda_i)}} \Bigg]^{\zeta_i},
\end{aligned}
\end{equation}
where $\beta_i, \zeta_i > 0$.
Considering that all capability dimensions must be assessed concurrently \textcolor{black}{and assuming that the capability dimensions are represented by independent random variables}, for the probability computation, we have
\begin{equation}
\label{eq:trustGivenCapability_all}
\begin{aligned}
    p\big( \Omega(\lambda, \bar{\lambda}) = 1 | \lambda \big) = \prod_{i=1}^n \tau_i = \prod_{i=1}^n \Bigg[ \frac{1}{1 + e^{\beta_i (\bar{\lambda}_i - \lambda_i)}} \Bigg]^{\zeta_i},
\end{aligned}
\end{equation}
where $t$ was suppressed, as the resulting function is independent of the time.
\textcolor{black}{The product of probabilities in Eq. (\ref{eq:trustGivenCapability_all}) can quickly converge to zero as $n$ increases. 
Therefore, to improve code implementation stability in practical implementations, a linear form of Eq. (\ref{eq:trustGivenCapability_all}) could be used (i.e., by taking the logarithm on both sides of the equation}).

\begin{figure*}
  \centering
  \includegraphics[width=1.0\linewidth]{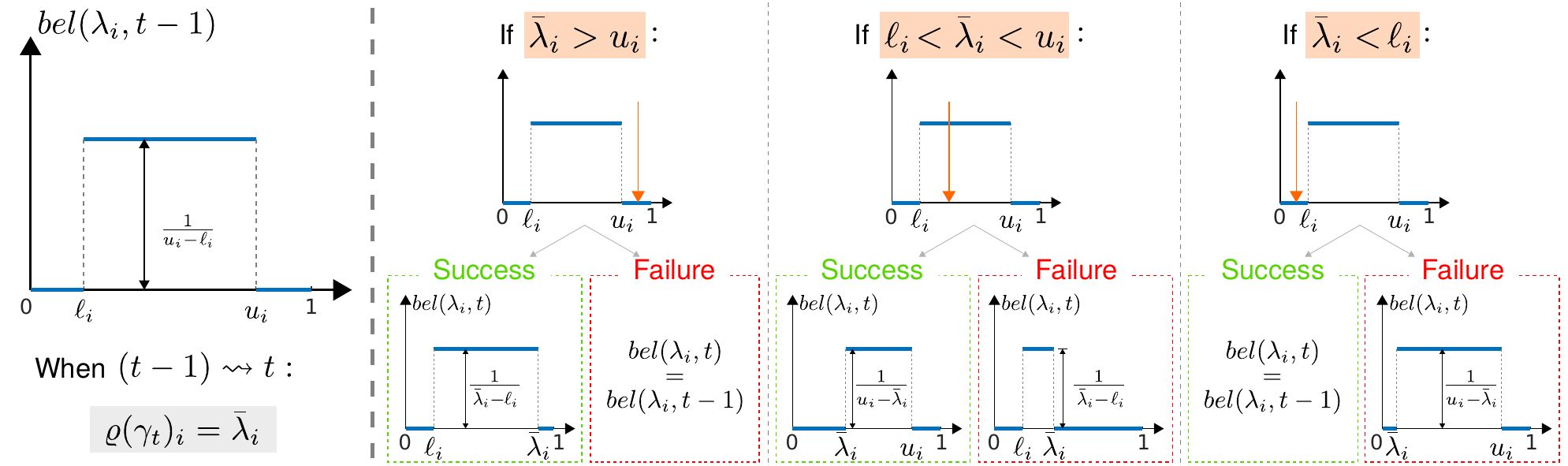}
  \caption{
  Capability update procedure, where each capability dimension changes after the trustor agent observes the trustee agent $a$ executing a task $\gamma_t$ (at a specific time instance $t$).
  The belief distribution over $a$'s capabilities \textit{before} the task execution $bel(\lambda_i, t-1)$ is updated to $bel(\lambda_i, t)$, depending on the task capability requirements $\varrho(\gamma_t)_i = \bar{\lambda}_i$ and on the performance of $a$ in $\gamma_t$, which can be a success ($\Omega = 1$) or a failure ($\Omega = 0$).
  \textcolor{black}{The capability belief: (i) expands either when the agent succeeds on a task whose requirement exceeds $u_i$, or when the agent fails on a task whose requirement is less than $\ell_i$; (ii) contracts when the agent succeeds or fails on a task whose requirement falls between $u_i$ and $\ell_i$; or (iii) remains the same either when the agent fails on a task whose requirement exceeds $u_i$, or when the agent succeeds on a task whose requirement is less than $\ell_i$.
  }
  }
  \label{fig:algorithm}
\end{figure*}

Trust dynamics is established with a process for updating $bel(\lambda, t)$ that relates observations of a trustee agent’s past performances with that agent’s likelihood of success on related tasks.
We considered that a trustor agent must build the belief about the trustee's capabilities after observations of the trustee's performances.
However, initially, the trustor has no information about the trustee's performances and capabilities.
We assumed this is represented by $bel(\lambda, 0)$ being a uniform probability distribution over the capability hypercube $\Lambda$, i.e., 
$bel(\lambda_i, 0) = \mathcal{U}(0, 1), \forall i \in \{1,2, ..., n\}$.
Next, after observing the sequence of successes and failures of the trustee in different tasks, the trustor updates $bel(\lambda, t)$, following the procedures in Algorithm \ref{alg:update_bel} and in Fig. \ref{fig:algorithm}

\begin{algorithm}
    \caption{
    Capability Belief Initialization and Update
    }
    \label{alg:update_bel}
    \begin{algorithmic}[1]
    
    \Procedure{Capability Hypercube Initialization}{}
    
    \For{$i = 1:n$}
    
    \State $\ell_i \gets 0$
    
    \State $u_i \gets 1$
    
    \State $bel(\lambda_i, 0) \gets \mathcal{U}(\ell_i,  u_i)$ \Comment{\textit{Uniform distributions}}
    
    \EndFor
    
    \EndProcedure
    
        \Procedure{Capability Update}{$\gamma, bel(\lambda, t-1)$}
        
    \Comment{\textit{When trustor observes trustee executing $\gamma$ at $t$}}
        
        \For{$i = 1:n$}
            \If{$\Omega(\lambda, \bar{\lambda}, t) = 1$}
                \If{$\bar{\lambda}_i > u_i$}
                    \State $u_i \gets \bar{\lambda}_i$
                \ElsIf{$\bar{\lambda}_i > \ell_i$}
                    \State $\ell_i \gets \bar{\lambda}_i$
                \EndIf
            \ElsIf{{$\Omega(\lambda, \bar{\lambda}, t) = 0$}}
                \If{$\bar{\lambda}_i < \ell_i$}
                    \State $\ell_i \gets \bar{\lambda}_i$
                \ElsIf{$\bar{\lambda}_i < u_i$}
                    \State $u_i \gets \bar{\lambda}_i$
                \EndIf
            \EndIf
            \State $bel(\lambda_i, t) \gets \mathcal{U}(\ell_i, u_i)$
        \EndFor
        \EndProcedure
    \end{algorithmic}
\end{algorithm}

\subsection{Artificial Trust}

For representing the artificial trust of a robotic trustor in a trustee agent, the bi-directional trust model can be slightly modified.
We can vanish subjective biases that characterize human trustors by considering large values for the parameters $\beta_i$ in Eq. (\ref{eq:trustGivenCapability_dim}) (i.e., considering the robot to be ``infinitely pragmatic'').
With sufficiently large $\beta_i$, $\tau_i$ becomes an analytic approximation of a decreasing step function with the transition from $1$ to $0$ when $\bar{\lambda}_i = \lambda_i$, i.e.
\begin{equation}
\label{eq:trust_heaviside}
    \begin{aligned}
        \lim_{\beta_i \to \infty} \tau_i = \mathcal{H}\big(-\bar{\lambda}_i + \lambda_i\big),
    \end{aligned}
\end{equation}
where $\mathcal{H}(x)$ is the Heaviside function of a continuous real variable $x$.
Considering all capability dimensions to be independent, and using the approximation in Eq. (\ref{eq:trust_heaviside}) for computing trust with Eq. (\ref{eq:trustGivenCapability_all}) and Eq. (\ref{eq:trust}), we have
\begin{equation}
\label{eq:artificial_trust_product}
    \begin{aligned}
        \tau(a, \gamma, t) = \prod_{i=1}^n \psi(\bar{\lambda}_i),
    \end{aligned}
\end{equation}
where,
\begin{equation}
\label{eq:artificial_trust}
    \begin{aligned}
        \psi(\bar{\lambda}_i) = 
        \left\{
        	\begin{array}{ll}
        		1  & \mbox{if } 0 \leq \bar{\lambda}_i \leq \ell_i, \\
        		\frac{u_i - \bar{\lambda}_i}{u_i - \ell_i}  & \mbox{if } \ell_i < \bar{\lambda}_i < u_i, \\
        		0  & \mbox{if } u_i \leq \bar{\lambda}_i \leq 1.
        	\end{array}
        \right.
    \end{aligned}
\end{equation}

For each capability dimension, the robotic trustor agent believes that the trustee agent's capability is a random variable $\lambda_i$ uniformly distributed between $\ell_i$ and $u_i$.
If a task requires $\bar{\lambda}_i <  \ell_i$, the trustee capability exceeds the task requirement and trust is $1$.
Conversely, if $\bar{\lambda}_i > u_i$, the task requirement exceeds the trustee's capability and trust is $0$.
In the intermediate condition, trust decreases with a constant slope from $1$ to $0$, corresponding to $\bar{\lambda}_i = \ell_i$ and $\bar{\lambda}_i = u_i$.

Robots can use long-term information to update their capability beliefs with a process different from that presented in Algorithm \ref{alg:update_bel}.
An alternative is to recursively solve an optimization problem, considering the history of outcomes observed from different tasks $\gamma$ (with different $\varrho(\gamma) = \bar{\lambda} \in \Lambda$).
Trust is approximated by the number of successes divided by the number of times the task $\gamma$ was performed, i.e.,
\begin{equation}
\label{eq:computed_artificial_trust}
    \begin{aligned}
        \hat{\tau} = \frac{\sum\limits_{m=0}^t \Omega(\xi(a), \varrho(\gamma), m)}{\sum\limits_{m=0}^t \big[\Omega(\xi(a), \varrho(\gamma), m) + \mho(\xi(a), \varrho(\gamma), m)\big]},
    \end{aligned}
\end{equation}
and, considering each $\lambda = \varrho(\gamma)$, the capability distribution limits $\ell_i$ and $u_i$ should be chosen such that $bel(\lambda, t) = \prod_{i=1}^n \mathcal{U}(\hat{\ell}_i, \hat{u}_i)$, and
\begin{equation}
\label{eq:optimization_artificial_trust}
    \begin{aligned}
        (\hat{\ell}_i, \hat{u}_i) = \argmin_{[0, 1]^2} \int_{\Lambda} \| \tau - \hat{\tau} \|^2 d\lambda.
    \end{aligned}
\end{equation}

For numerical computations, $\Lambda$ can be discretized and Eq.  (\ref{eq:optimization_artificial_trust}) approximated with a summation, as in Section \ref{ssec:robot_trust}.

\section{Experiment}
\label{sec:experiment}

We conducted an online experiment using a Qualtrics survey and the Amazon Mechanical Turk (MTurk) platform to gather a dataset for comparing our model with other trust prediction models, such as Soh's models \cite{Soh2020Multi-taskInteraction} and OPTIMo \cite{Xu2015OPTIMo:Collaborations}.
We aimed to emulate a human-automated vehicle (AV) interaction setting, asking participants to (1) assess the requirement levels for driving tasks that were to be executed by the AV, (2) watch videos of the AV executing a part of those tasks and (3) evaluate their trust in the AV to execute other tasks (distinct from those they have watched in the videos).

\begin{figure*}
  \centering
  \includegraphics[width=1.0\linewidth]{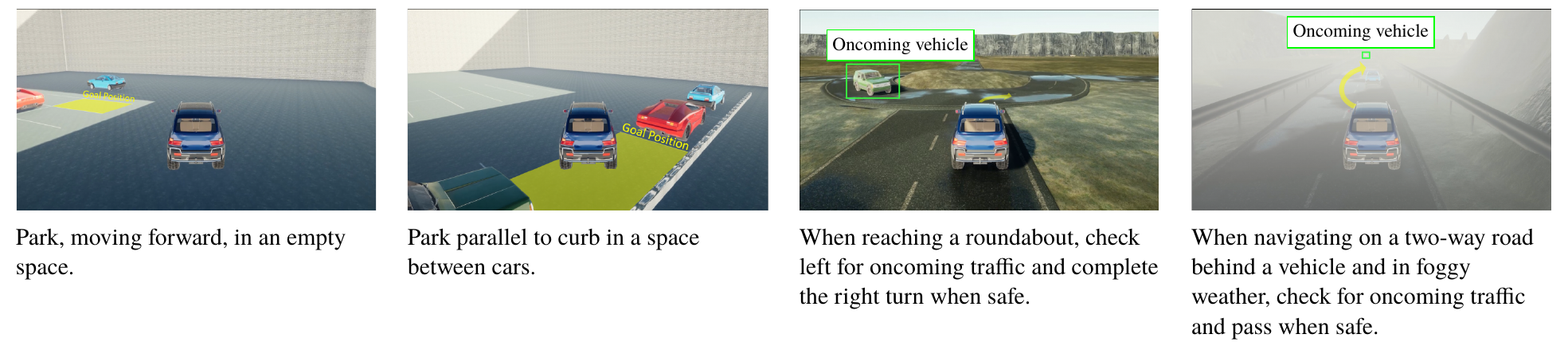}
  \caption{
  Tasks presented to the experiment participants in terms of images and corresponding verbal descriptions.
  The participants had to rate the capability requirements for each of these tasks, considering two capability dimensions: sensing and processing.
  In other words, they had to assign a pair $(\bar{\lambda}_1, \bar{\lambda}_2) \in [0, 1]^2$ for each task.
  Tasks were randomly presented for avoiding ordering effects.
  }
  \label{fig:tasks_figs_descriptions}
\end{figure*}

Initially, only images and verbal descriptions of four driving tasks were presented in random order to the participants (Fig. \ref{fig:tasks_figs_descriptions}).
Participants were asked to rate the capability requirements for each of the presented tasks in terms of two distinct capabilities of the AV: sensing and processing, which were defined and presented to the participants as,

\begin{itemize}
    \item \textbf{Sensing ($\lambda_s$) -} \textit{The accuracy and precision of the sensors used to map the environment where the AV is located and perceive elements within that environment, such as other vehicles, people and traffic signs.}
    
    \item \textbf{Processing ($\lambda_p$) -} \textit{The speed and performance of the AV’s computers that use the information from sensors to calculate the trajectories and the steering, acceleration, and braking needed to execute those trajectories.}

\end{itemize}

Participants were asked to indicate the required capability levels $(\bar{\lambda}_s, \bar{\lambda}_p) \in [0, 1]^2$ for each task, providing a score (i.e., indicating a slider position on a continuous scale) varying from low to high.

After evaluating all four presented tasks, participants watched short videos (approximately $20s$ to $30s$) of a simulated AV executing three of the four tasks.
Those three were considered \textit{observation tasks}.
The videos showed the AV succeeding or failing to execute each observation task.
(All videos are available at \url{https://bit.ly/37gXXkI}.)
Next, participants were asked to indicate whether the AV successfully executed the task.
That question served both as an attention checker and as a way to make the participant acknowledge the performance of the AV in that specific task.
After watching each video, participants were also asked to rate their trust $\tau$ in the AV to execute the fourth remaining task (i.e., the \textit{trust prediction task}) on a 7-point Likert scale varying from ``very low trust'' to ``very high trust'', as an indication of how much they disagreed or agreed with the sentence: ``\textit{I believe that the AV would successfully execute the task}.''
Participants were asked to consider all videos they had seen during the observation tasks and rate their trust in the AV to execute the trust prediction task.
Finally, participants received a random 4-digit identifier code to upload in the MTurk platform and receive their payment.

To keep work-related regulations consistent, we restricted our participants to those who were physically in the United States when accepting the MTurk human intelligence task (HIT).
A total of 284 MTurk workers participated in our experiment and received a payment of \$1.80 for completing the HIT without failing to correctly answer the attention checker questions.
The HITs were completed in approximately 6min40s, on average.
We collected no demographics data or other personal information from the participants because these were not needed for our analyses.
The obtained dataset and our implementations are available at \url{https://bit.ly/3sfVtuK}.
The research was reviewed and approved by the University of Michigan’s institutional review board (IRB\# HUM00192470).

\section{Results}
\label{sec:results}

\subsection{Human-drivers' (natural) trust in robotic AVs}
\label{ssec:human_trust}

We implemented a $10$-fold cross-validation to train and evaluate our bi-directional trust model (BTM) with the data obtained in the experiment described in Section \ref{sec:experiment}.
For comparison, we also evaluated the performance of Soh's Bayesian Gaussian process model (GP) \cite{Soh2020Multi-taskInteraction} and that of a linear Gaussian model similar to Xu and Dudek's OPTIMo (OPT) \cite{Xu2015OPTIMo:Collaborations} on our collected dataset.
We obtained the tasks' vector representations for the GP model with GloVe \cite{pennington2014glove}, by processing the verbal descriptions presented in Fig.  \ref{fig:tasks_figs_descriptions}.
There were no closed forms for Eq. (\ref{eq:trust}), therefore we discretized each task capability dimensions in 10 equal parts and computed numerical approximations for $\tau$.
Because we considered only two outcome possibilities (failure or success in executing a task), the trust measurements from both the dataset and the model outputs were considered probability parameters of Bernoulli distributions. 
We considered the cross entropy between those distributions to be the loss function to be minimized.
We used PyTorch \cite{NEURIPS2019_9015} to implement all parameter optimizations with the Adam algorithm \cite{kingma2014adam}, using randomized validation sets comprising $15\%$ of the training data.
Two metric scores were computed for the comparisons among model performances: the mean absolute error (MAE); and the negative log-likelihood (NLL), which corresponds to the loss function chosen for the optimizations.

\begin{figure}[ht]
  \centering
  \includegraphics[width=1.0\linewidth]{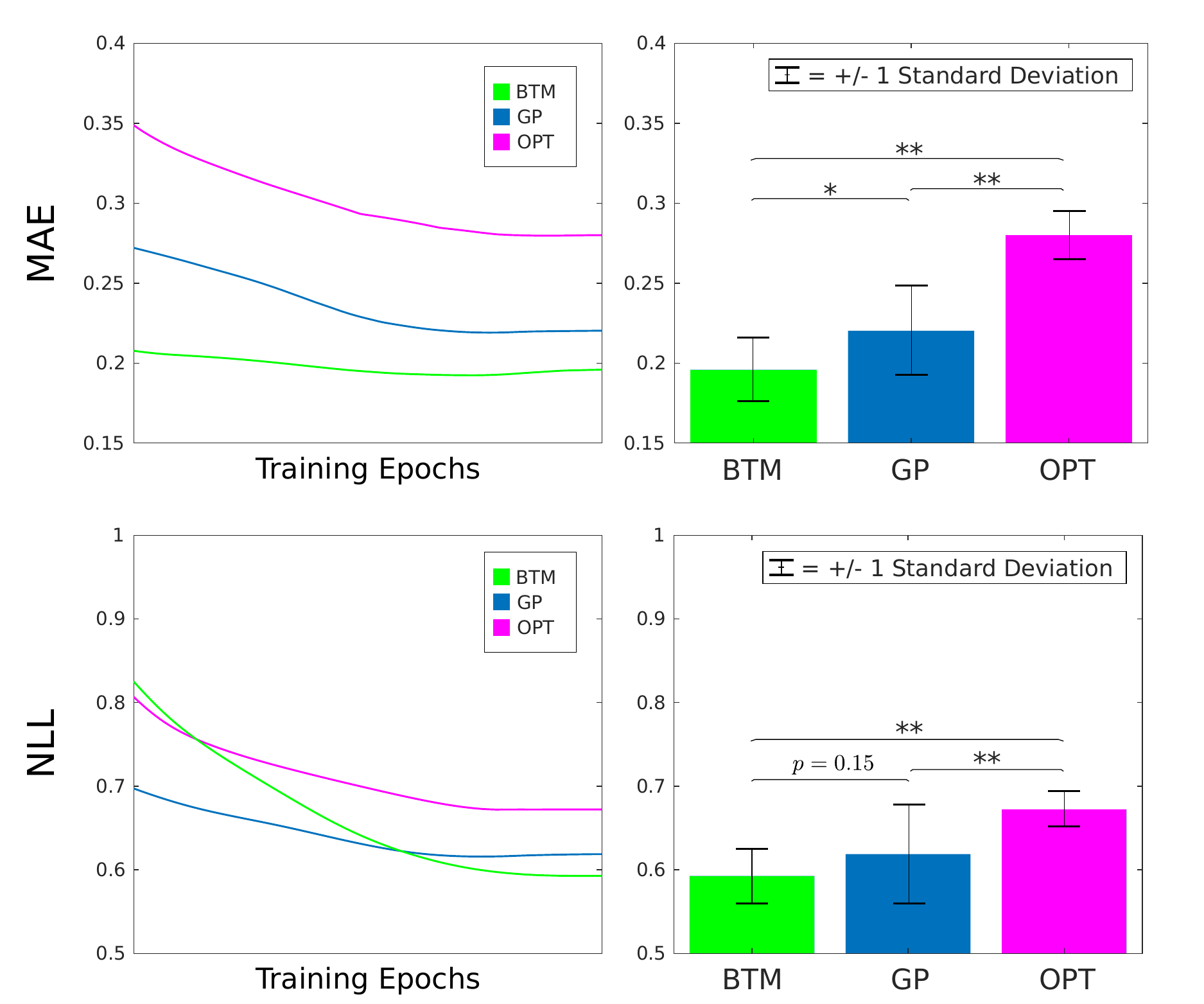}
  \caption{
  MAE and NLL learning curves and final values for our proposed trust model (BTM) and for current trust models from \cite{Soh2020Multi-taskInteraction} (GP) and \cite{Xu2015OPTIMo:Collaborations} (OPT).
  As the total number of training epochs is different for each model, their representation on the horizontal axes of the learning curves is normalized.*$p<0.05$; **$p<0.01$.
  }
  \label{fig:results_human_trust}
\end{figure}

Table \ref{tab:MAE_NLL_results} presents the MAE and NLL scores averaged over the $10$ cross-validation folds (with standard deviations between parentheses) for the BTM, GP and OPT models.
Fig. \ref{fig:results_human_trust} complements the table, showing the average learning curves for both scores and bars representing the average final values with $\pm 1$ standard deviations.

\begin{table}[ht]
\renewcommand{\arraystretch}{1.3}
\caption{
Mean Absolute Error (MAE) and Negative Log-Likelihood (NLL) average minimized scores for each trust model
}
\label{tab:MAE_NLL_results}
\centering
\begin{tabular}{c c c}
    \hline
    \textbf{Model}  &  \textbf{MAE}$^{\dagger}$ & \textbf{NLL}$^{\dagger}$\\
    \hline

    BTM   &   $\mathbf{0.196 (0.020)}^{\ddagger}$ & $\mathbf{0.593 (0.033)}^{\ddagger}$\\

    GP    &   $0.220 (0.028)$ & $0.619 (0.060)$\\
    
    OPT   &   $0.280 (0.016)$ & $0.672 (0.021)$\\
    \hline

\end{tabular}
    \begin{tablenotes}
      \small
      \item \hspace{0.9cm} $^\dagger$10-fold results: Mean(Standard Deviation).
      \item \hspace{0.9cm} $^\ddagger$Best scores in \textbf{bold}.
    \end{tablenotes}
\end{table}

Our bi-directional trust model (BTM) outperformed both the GP and the OPT models after the parameter optimization process.
BTM reduced the MAE metric by approximately $11\%$ as compared with GP, and by $30\%$ as compared to OPT.
In terms of NLL, the use of BTM reduced this metric by approximately $4.3\%$ as compared with GP model, and by $12\%$ as compared with the OPT model.

\subsection{Robots' Artificial Trust in Humans}
\label{ssec:robot_trust}

\begin{figure*}[ht]
  \centering
  \includegraphics[width=1.0\linewidth]{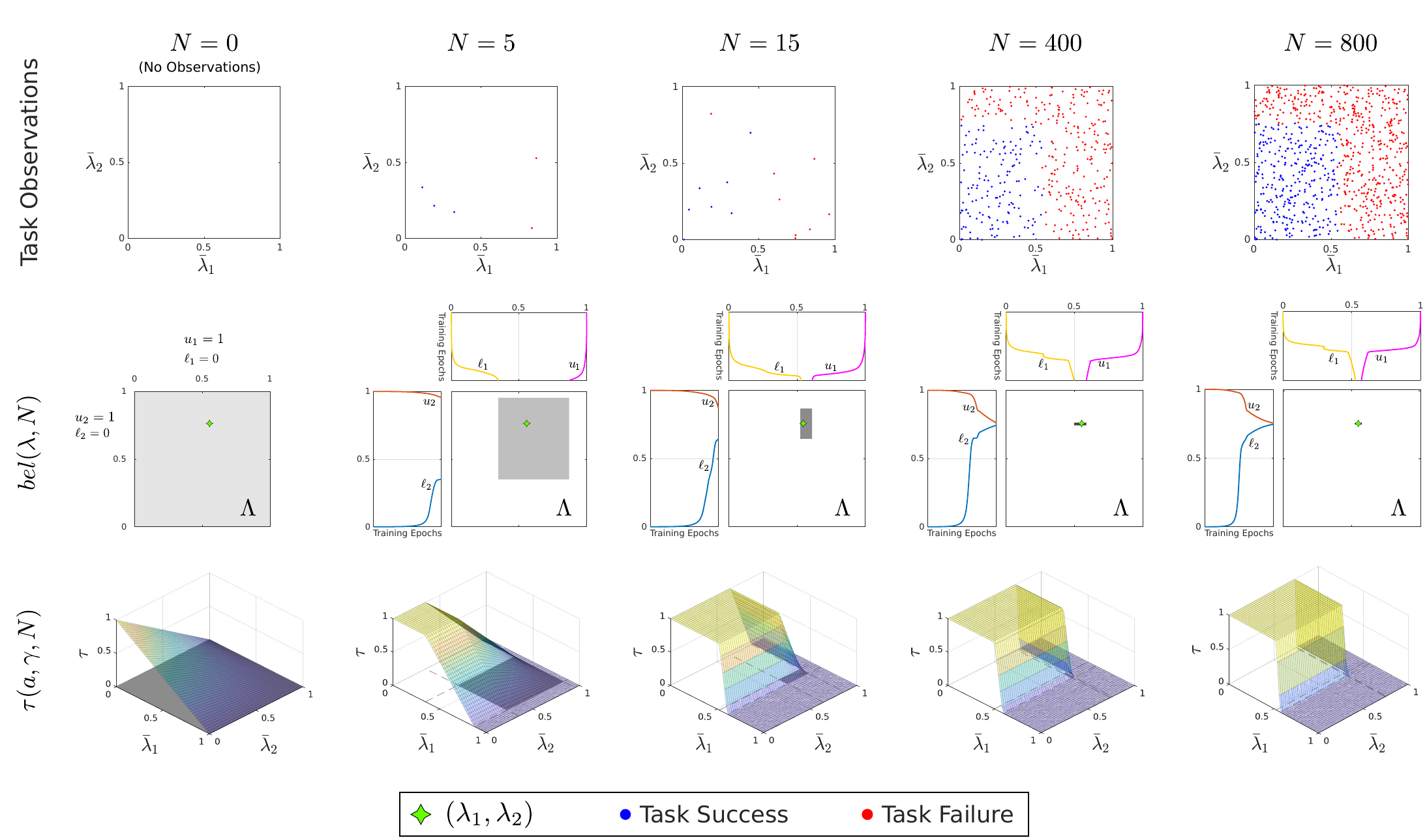}
  \caption{
  Artificial trust results, where a robotic trustor agent's belief over a trustee agent $a$'s capabilities is updated after $N$ observations of $a$'s performances in different tasks, represented by points in $\Lambda = [0, 1]^2$.
  When $N = 0$, $bel(\lambda, N)$ is ``spread'' over the entire $\Lambda$.
  When the robot trustor collects observations, it starts building $a$'s capabilities profile and reducing the gray area in the $bel(\lambda, N)$ distribution.
  This profile gets more accurate when $N$ increases and $(\lambda_1, \lambda_2)$ gets better defined.
  This is also reflected in the evolution of the conditional trust function $\tau(a, \gamma, N)$.
  }
  \label{fig:robot_trust_results}
\end{figure*}

Besides evaluating and comparing our bi-directional trust model with other trust models using experimental data, we also implemented simulations to verify its use in the artificial trust mode (i.e., as a model for predicting a robots' trust in another trustee agent).
We assumed two unspecified capability dimensions, considering that a trustee agent $a$'s capabilities were static and represented by a point $\xi(a) = (\lambda_1, \lambda_2) \in \Lambda = [0, 1]^2$.
The trustee agent's capabilities were initially unknown by the trustor robot, who must estimate $\xi(a)$ after observing the trustee's performances in several different tasks.
We considered $N$ fictitious tasks $\gamma^j$, $j \in \{1, 2, ..., N\}$, and randomly picked $N$ points $\varrho(\gamma^j) = (\bar{\lambda}_1^j, \bar{\lambda}_2^j) \in \Lambda$ representing capability requirements for the tasks.
Task outcomes were assigned to each of the $N$ tasks, with high probability of success for tasks that simultaneously had $\bar{\lambda}_1^j \leq \lambda_1$ and $\bar{\lambda}_2^j \leq \lambda_2$, and low probability of success when $\bar{\lambda}_1^j > \lambda_1$ or $\bar{\lambda}_2^j > \lambda_2$.
Again, for numerical computations, we discretized both capability dimensions in $10$ equal parts, obtaining $100$ bins for $\Lambda$.
We computed the observed probabilities of success for tasks inside a bin dividing the number of successes by the total number of tasks that fell on each bin (i.e., the approximation for $\hat{\tau}$).
Finally we ran optimizations to find the parameters that best characterized $bel(\lambda_1, N)$ and $bel(\lambda_2, N)$, solving the problem represented by Eq. (\ref{eq:optimization_artificial_trust}).
Fig. \ref{fig:robot_trust_results} illustrates the evolution of $bel(\lambda, N)$ and of $\tau(a, \gamma, N)$ for increasing values of $N$.
The higher the number of observations, the better the accuracy of $a$'s identified capabilities.

\section{Discussion}
\label{sec:discussion}

\textcolor{black}{Our model is based on general capability representations that can be either performance or non-performance trust factors.
This particular aspect of our bi-directional trust model makes it useful for representing a robot's artificial trust, as presented in Subsection \ref{ssec:robot_trust}, and allows for better human trust predictions in comparison to existing models, as presented in Subsection \ref{ssec:human_trust}.
Additionally, our model considers task capability requirements in its description, describing how hard a task is for an agent to execute.
The model's mathematical formulation captures the differences between those task requirements and the potential trustee agent's observed capabilities.
Differently from the Gaussian process-based method presented in \cite{Soh2020Multi-taskInteraction}, this formulation allows for the adequate representation of lower trust levels when the requirements of a task exceed the capabilities of the agent and, conversely, higher trust levels when the agent capabilities exceed the task requirements.}

The results reveal that our proposed bi-directional trust model has better performance for predicting a human's trust in a robot (in our specific experiment, an AV) than the models from \cite{Xu2015OPTIMo:Collaborations} and \cite{Soh2020Multi-taskInteraction}.
This performance improvement was expected because current models are limited in capturing important trust-related parameters, such as the agents' capabilities or task's requirements in their formulation.
To the best of our knowledge, only our model and Soh's models \cite{Soh2020Multi-taskInteraction} distinguish and describe the trust transfer between different tasks, while OPTIMo \cite{Xu2015OPTIMo:Collaborations} is more appropriate for predicting a human's trust in a robot to execute one specific task.

\textcolor{black}{Section \ref{ssec:robot_trust} presents simulations that show how the proposed model can be used for representing a robot's artificial trust.
In the future, the proposed bi-directional trust model could be used in real-world human subjects experiments.
An example could be a study where participants would execute some tasks represented in the capabilities hypercube, and the robot would be able to establish its trust in the participants based on their failures or successes on those tasks.
In parallel, the robot could estimate the human's natural trust for different tasks, and use both natural and artificial trust metrics to compute expected rewards for the execution of new tasks.
Tasks could be allocated between the human and the robot to maximize the expected reward of a whole set of tasks, eventually improving the joint performance of the human--robot team.}

Despite the eventual improvement on multi-task trust prediction performance, the use of task capability requirements could also be considered a drawback of our model because it calls for one more subjective input dimension in comparison with current models.
Rating and describing tasks that must be executed by humans and robots in terms of specific human/robotic capability dimensions depends on the trustor agent's individual beliefs and experiences---natural, in the case of a human trustor agent, or artificial, in the case of a robotic trustor agent.
Our models' trust prediction performance might have also been restricted by inconsistencies related to task characterization by each participant of our experiment.
We believe that better trust prediction results can be achieved with in-person longitudinal experiments involving fewer participants and more predictions.

\section{Conclusion}
\label{sec:conclusion}

We presented a multi-task bi-directional trust model that depends on both a trustee agent's proven capabilities (as observed by the trustor agent) and on the task capability requirements (as characterized by that same trustor agent).
Our model outperformed the most relevant and recent trust models (i.e., \cite{Xu2015OPTIMo:Collaborations} and \cite{Soh2020Multi-taskInteraction}) in terms of predicting the transferred trust between distinct tasks by addressing the main limitations of those models.
With a generalist capability dimension representing trustee agents' capabilities, our model can also represent robots' artificial trust in different trustee agents.
Our model is useful for future applications where humans and robots collaborate and must sequentially take turns in executing different tasks.

\bibliographystyle{IEEEtran}
\bibliography{Mendeley_Refs}

\end{document}